\documentclass{article}

\PassOptionsToPackage{numbers}{natbib}

\usepackage[main, final]{neurips_2026}

\usepackage[utf8]{inputenc} 
\usepackage[T1]{fontenc}    
\usepackage{hyperref}       
\usepackage{url}            
\usepackage{booktabs}       
\usepackage{amsfonts}       
\usepackage{nicefrac}       
\usepackage{microtype}      
\usepackage{xcolor}         
\usepackage{wrapfig}
\usepackage{graphicx}
\usepackage{xcolor}
\usepackage{xspace}
\usepackage{amsmath}
\usepackage{subcaption}
\usepackage{array}
\usepackage{multirow} 
\usepackage{wrapfig}
\definecolor{darkgreen}{RGB}{0,100,0}
\usepackage{enumitem}
\usepackage{xfrac}
\usepackage{float}

\title{HiFloat4 Format for End-To-End Reinforcement Learning Post-Training of Large Language Models}

\newcommand{\ourmethod}{Rollout-ResQ\xspace}
\newcommand{\ourmethodc}{\ourmethod-$S_{50\%}$}
\newcommand{\ourmethods}{\ourmethod-$S_{2:4}$\xspace}
\newcommand{\ourmethodb}{\ourmethod-$S^{32\times 64}_{50\%}$\xspace}
\newcommand{\ourmethodd}{\ourmethod-Dense\xspace}

\newcommand{\review}[1]{{\color{black}#1}}

\author{%
  \And
  Hei Yi Mak\\
  \And
  Shadan Golestan\thanks{indicates equal contribution. These authors are ordered alphabetically by last name. }\\
  \And
  Hoang Le\footnotemark[1]\\
  \And
  Mehran Taghian\footnotemark[1]\\
  \And
  Yunke Peng\\
  \And
  Yaoyuan Wang\\
  \And
  Yao Wang\\
  \And
  Junsong Wang\\
  \And
  Tianchi Hu\\
  \And
  Fengchen He\\
  \And
  Guipeng Hu\\
  \And
  Tanzila Rahman\\
  \And
  Anandharaju Durai Raju\\
  \AND
  \textbf{Huawei}\\
  \texttt{\{hei.yi.mak1, pengyunke\}@huawei.com}
}

\begin{document}

\maketitle

\begin{abstract}
  We present, to our knowledge, the first  \emph{end-to-end} FP4 RL post-training, 
    in which both the rollout and training policies, 
    including their forward and backward passes, operate at 4-bit precision.
    A systematic study reveals that the dominant source of degradation in FP4 RL is not training-side quantization error but rollout activation quantization: 
    outliers stretch the dynamic range so far that a large number of activation values 
    underflow to zero under FP4. 
    Counterintuitively, restoring the training policy to higher precision 
    while keeping the rollout in FP4 makes accuracy \emph{worse} than full FP4 baseline, 
    exposing rollout--training mismatch as the principal failure mode 
    and ruling out standard pretraining-style fixes. 
    We address this with \textbf{Rollout} \textbf{Res}idual \textbf{Q}uantization (\textbf{\ourmethod}): 
    a single residual correction term constrained to a hardware-friendly sparsity pattern, 
    added only to the FP4 rollout matmul --- a lightweight correction 
    that recovers most of the precision lost to outlier-driven underflow 
    without inflating the rollout's compute footprint.
    \review{ On Qwen2.5-3B}, \ourmethod paired with the HiFloat4 (HiF4) format ---
    whose three-level hierarchical scaling preserves resolution under FP4's tight 4-bit budget ---
    closes the accuracy gap to BF16 from 
    \review{$3.59\%$ to $1.39\%$.}
    bringing fully quantized FP4 RL within \review{close reach} of full precision. 
    Applied to the open-standard MXFP4, the same recipe narrows the gap from $13.6\%$ to $5.3\%$, revealing that FP4 format choice is a key factor that determines the ceiling on recoverable accuracy.
    The experimental results on Qwen2.5-Math-7B further confirm the recovery benefits of \ourmethod with HiF4.
    Together, these results establish HiF4 as the enabling format 
    for end-to-end FP4 RL post-training, 
    and \ourmethod as the activation-side mechanism that makes the gap to BF16 closable. 
\end{abstract}

\section{Introduction}

Large Language Models (LLMs) have achieved remarkable performance 
across a wide range of applications~\cite{brown2020language,10.5555/3600270.3602281,laskar2024systematic}.
By leveraging extensive Chain-of-Thought reasoning, 
LLMs have demonstrated strong capabilities in challenging domains~\cite{NEURIPS2022_9d560961, zhou2024selfdiscover} 
such as mathematical problem solving~\cite{dinucu2025problem}, 
code generation~\cite{zhao2025qspec}, 
and agentic workflows~\cite{wei2025webagent,bo2024reflective}.
In doing so, these models produce long reasoning trajectories
that interleave exploration, verification, and self-correction.
Eliciting such behavior at scale has driven the widespread adoption of 
Reinforcement Learning (RL) for LLM post-training~\cite{liao2025enhancing,zhang2025rearank,liu2025nover,xie2025logic}, 
most notably in the form of RL with Verifiable Rewards (RLVR) 
using policy-gradient algorithms such as Group Relative Policy Optimization (GRPO)~\cite{shao2024deepseekmath} 
and its variants~\cite{zhang2025grpo,liu2025understanding,liu2026gdpo}.

Despite its success, RL post-training incurs substantial computational, 
memory, and monetary costs~\cite{hu-etal-2025-openrlhf,liao2025enhancing,song2025fastcurl}. 
Each iteration consists of two phases: 
a \emph{rollout phase}, in which a \emph{rollout policy} samples responses for each prompt, 
and a \emph{training phase}, in which a \emph{training policy} evaluates those responses and is updated. 
The rollout phase is typically the dominant bottleneck, 
since responses must be sampled autoregressively over long horizons~\cite{liao2025enhancing,jetrl,hadji-kyriacou2024would}---
a cost that is amplified in reasoning tasks, 
where longer trajectories provide stronger learning signals~\cite{jetrl}. 
A natural remedy is low-precision computation, 
in particular quantization~\cite{courbariaux2016binarized}, 
which reduces memory usage and accelerates 
the matrix multiplications that dominate both generation and the forward/backward passes. 
Quantizing the rollout policy addresses this bottleneck, 
but quantizing the training policy is equally important for end-to-end efficiency~\cite{jetrl}: 
beyond accelerating the forward and backward passes, 
it narrows the numerical mismatch between the two policies, 
yielding more stable post-training.

To improve the efficiency of RL post-training, prior work has explored 
low-precision computation with 8-bit formats such as FP8 or INT8, 
applied either to the rollout policy alone~\cite{qurl} 
or to both the rollout and training policies~\cite{jetrl}. 
A central concern in these methods is the numerical mismatch 
between a quantized rollout policy and a higher-precision training policy, 
which they address through importance-sampling correction~\cite{qurl} 
or a unified low-precision pipeline~\cite{jetrl}. 
More recently, QeRL~\cite{qerl} has pushed the rollout policy to 4-bit (NVFP4) 
while keeping gradients in higher precision through LoRA adapters, 
demonstrating that 4-bit rollouts are feasible without destabilizing training. 
Together, these results show that reduced precision 
can substantially accelerate training and lower memory overhead.
In this work, we go a step further and study \emph{end-to-end} FP4 RL post-training, 
in which both the rollout and training policies---
including their forward and backward passes---operate in 4-bit. 
This regime is considerably more challenging than the settings above: 
unlike QeRL, we do not retain a higher-precision gradient path, 
and unlike 8-bit methods, we operate at a precision where the limited numerical 
range and resolution can destabilize optimization, 
yielding fragile and often non-convergent training dynamics.

We build on HiFloat4 (HiF4)~\citep{luo2026hifloat4}, 
which uses a three-level hierarchical scaling scheme 
to expand dynamic range while preserving precision within FP4's tight 4-bit budget. 
We additionally evaluate against MXFP4~\citep{rouhani2023mxfp4}, 
the open-standard 4-bit format 
with a single shared scale per block, 
to test whether our findings depend on HiF4 specifically. 
Prior work~\cite{taghian2026hifloat4formatlanguagemodel}
showed that HiF4 enables fully quantized FP4 pretraining
with only minimal precision-recovery techniques on the training pass, 
matching full-precision baselines. 
The RL post-training setting, however, shifts where the difficulty lies, 
and whether the pretraining recipe transfers---and, if not, where the new difficulty resides---
is the question we take up next.

We address this question through a systematic study 
of where precision actually matters in FP4 RL post-training. 
We localize the dominant source of degradation 
to rollout-side \emph{activation} quantization, 
rather than to the training-pass quantization. 
Strikingly, restoring the training policy to higher precision 
while keeping the rollout in FP4 does not just fail to recover accuracy---
it performs \emph{worse} than fully FP4 training, 
suggesting that the rollout--training mismatch, not training-side noise, 
is the principal failure mode. 
To understand the rollout-side problem, 
we inspect the activation distributions 
and find that outliers stretch the dynamic range so far 
that the bulk of activation values underflow to zero under FP4. 
Preliminary experiments reveal that keeping the rollout activations in higher-precision representation 
recovers most of the accuracy lost relative to the baseline---
but this fix forfeits the very benefit we set out to capture, 
since mixed-precision GEMMs (higher-precision activations $\times$ FP4 weights) 
give up the throughput of true FP4 matmul.

A natural next step is to draw on the post-training quantization (PTQ) literature, 
which has developed a rich toolkit for handling activation outliers: 
smoothing~\cite{xiao2023smoothquant}, scaling~\cite{lin2024awq}, 
clamping~\cite{wang2025optimizing}, learnable calibration transforms~\cite{shao2024omniquant}, 
and rotation- or permutation-based methods~\cite{ashkboos2024quarot,liu2025spinquant,lin2024duquant}. 
These methods, however, all assume a \emph{static} inference setting: 
activation statistics are estimated once on a calibration set, 
quantization parameters are fit offline, 
and the resulting transforms are reused unchanged at deployment. 
RL post-training violates each of these assumptions. 
As the policy is updated, the rollout distribution---
and with it the activation statistics---shifts continuously, 
so any calibration done at step $t$ is partially stale by step $t+1$, 
and recalibrating frequently to track this drift 
would erode the very efficiency gains motivating FP4 in the first place.
We therefore take a different route, 
adopting a per-tensor fallback mechanism inspired by~\cite{zhang2025accurate}, where outliers are separated from the activation tensor, 
resulting in two FP4 tensors to fully take advantage of low-precision GEMM. 
This recovers most of the gap to the BF16 baseline 
and, unlike the higher-precision activation alternatives, 
keeps the rollout in 4-bit end-to-end. 
The catch is that splitting the activation into two FP4 tensors 
doubles the number of FP4 GEMMs in the rollout, 
trading one form of inefficiency for another.

To remove this overhead while preserving the accuracy gain, 
we propose \textbf{Rollout} \textbf{Res}idual \textbf{Q}uantization (\textbf{\ourmethod}), 
which augments the standard FP4 rollout computation with a residual correction term 
constrained to a hardware-friendly sparsity pattern~\cite{mishra2021acceleratingsparsedeepneural}. 
The two ingredients are inseparable: 
the residual, inspired by the decomposition of~\cite{zhang2025accurate}, 
captures the outlier-driven error that vanilla FP4 rollouts incur, 
while the imposed sparsity ensures the correction itself 
runs at a small fraction of a dense FP4 GEMM. 
The result is a single, lightweight correction added to the FP4 rollout path---
preserving accuracy close to the higher-precision-activation alternatives 
while retaining true FP4 GEMM throughput end-to-end.
Our contributions are as follows:
\begin{itemize}[itemsep=2pt, parsep=0pt, topsep=2pt]
    \item \textbf{First end-to-end FP4 RL post-training.} 
    To our knowledge, we present the first end-to-end 
    FP4 RL post-training in which both the rollout and training policies---
    including their forward and backward passes---operate at 4-bit precision, 
    without LoRA adapters or higher-precision gradient paths.
    
    \item \textbf{Diagnosis.} 
    We localize the dominant source of degradation 
    to rollout-side \emph{activation} quantization, 
    where outliers cause the bulk of activation values to underflow under FP4. 
    Restoring only the training policy to higher precision 
    performs \emph{worse} than uniform FP4 training, 
    indicating that the rollout--training mismatch, 
    not training-side noise, is the principal failure mode.
    
    \item \textbf{\ourmethod.} 
    A single residual correction with a hardware-friendly sparsity pattern, 
    added to the FP4 rollout matmul. 
    The correction recovers most of the accuracy lost to outlier-driven underflow 
    while keeping the rollout path in FP4 GEMMs.
    
    \item \textbf{Results.} 
    On Qwen2.5-3B and Qwen2.5-Math-7B, \ourmethod with HiF4 
    closes the BF16 accuracy gap from $4.9\%$ to $1.1\%$; 
    with MXFP4, from $13.6\%$ to $5.3\%$. 
    The method is format-agnostic, 
    and HiF4 is the superior format for extreme-precision training.

\end{itemize}

\section{Related Work}

\paragraph{Quantization-aware training (QAT) for RL.}
Prior QAT methods for low-precision RL post-training have mainly framed the problem as a train-rollout policy mismatch, since the rollout policy is quantized (e.g., FP8) while the training policy remains in higher precision (e.g., BF16), addressing this through truncated importance sampling~\cite{yao2025on,liu2025flashrl}, adaptive quantization noise~\cite{qerl}, adaptive clipping, and update-aware quantization~\cite{qurl}. JetRL~\cite{jetrl} eliminates the mismatch by enforcing a unified low-precision pipeline across both policies. However, none of these methods account for the specific challenges of rollout activation quantization, which our results identify as the key bottleneck.

\paragraph{Scaling-based Methods.}

A rich body of work
addresses outliers in low-precision quantization 
through mixed-precision protection~\cite{dettmers2022gpt3,lee2024owq,kim2024squeezellm,dettmers2024spqr}, 
where a small subset of salient channels or weights 
is preserved in higher precision 
to reduce quantization error.
AWQ~\cite{lin2024awq} uses activation-aware scaling to protect salient weight channels, SmoothQuant~\cite{xiao2023smoothquant} migrates quantization difficulty from activations to weights via an equivalent transformation, and OmniQuant~\cite{shao2024omniquant} further introduces learnable weight clipping and equivalent transformations. All of these assume a frozen model with offline calibration data, making them ill-suited for RL post-training where activation statistics shift continuously as the policy evolves.

\paragraph{Rotation-based Methods.}
QuaRot~\cite{ashkboos2024quarot}, SpinQuant~\cite{liu2025spinquant}, DuQuant~\cite{lin2024duquant}, and~\cite{tseng2025trainingllmsmxfp4} redistribute extreme activation values across channels using rotation or random Hadamard transforms (RHT) to ease quantization. Like scaling-based methods, these transforms are typically fixed or optimized offline for a frozen model, making them less natural for RL post-training's evolving policy and non-iid rollout distribution.

\paragraph{Residual-based Methods.}

ARCQuant~\cite{meng2026arcquant} compensates quantization error through augmented residual channels, assuming a static setting where activation statistics can be estimated once. 
~\cite{wang2025optimizing} mitigates activation collapse via an outlier clamping-and-compensation (OCC) strategy, which does not require offline calibration and is thus more applicable to RL post-training. \cite{zhang2025accurate} propose a dynamic block-level fallback that augments the quantized activation with a residual term for outlier-detected blocks. Our method likewise separates the outlier-induced residual from the main activation term, but operates on a tensor level and replaces the dense fallback with a sparse residual correction, preserving most of the accuracy gain while substantially reducing FP4 GEMM overhead. Unlike dynamic block fallback, which targets static inference, we adapt a fallback mechanism to fully FP4 RL post-training where the correction must preserve rollout quality without sacrificing end-to-end low-precision efficiency.

\section{Preliminaries}\label{sec:3}
\paragraph{Block-Scaled Quantization}
Modern hardware-friendly 4-bit formats such as MXFP4~\cite{darvishrouhani2023microscaling} 
and HiF4~\cite{luo2026hifloat4} 
utilize block-scaled quantization to maintain numerical precision. 
Instead of a single per-tensor scale, 
a tensor is partitioned into blocks of size $B$ 
where the values in the same block share block or sub-block level scales 
to improve the representation capacity. 
MXFP4~\cite{darvishrouhani2023microscaling} uses block-wise microscaling 
over blocks of 32 elements 
to balance dynamic range and hardware efficiency. 
It employs an E8M0 format for the shared block-level scale 
and an FP4 E2M1 format for the individual elements. 
This design increases the representable dynamic range 
through a shared exponent per block, 
while keeping the per-element representation compact.
HiF4~\cite{luo2026hifloat4} uses a three-level hierarchical scaling scheme 
over blocks of 64 elements to balance dynamic range and numerical precision. 
Concretely, it employs an E6M2 format for the first-level scaler, 
E1 scalers for the second and third levels, 
and an S1P2 representation for the FP4 values, 
which is equivalent to E1M2. 
This hierarchical design increases the representable dynamic range 
while retaining precision through mantissa bits 
in both the top-level scaler and the FP4 values.

For a given linear layer $\ell$, 
let $X_{\ell} \in \mathbb{R}^{m \times d}$ denote the input activation matrix, 
$W_{\ell} \in \mathbb{R}^{n \times d}$ the weight matrix, 
and $Y_{\ell} = X_{\ell} W_{\ell}^\top$ the high-precision output, 
where $m, d, n$ represent 
the token count, 
input dimension, 
and output dimension, respectively. 
We denote a quantization configuration 
with $x$-bit weights and 
$y$-bit activations as $WxAy$. 
This work focuses on the $W4A4$ setting, 
where both weights and activations are quantized to FP4 format. 
Applying this to both the activations and the weights of a linear layer yields 
the low-precision approximation
\begin{equation}
\hat{Y}_{\ell} = \hat{X}_{\ell}\hat{W}_{\ell}^\top = Q(X_{\ell})Q(W_{\ell})^\top,
\end{equation}
where $Q$ is the quantization operator that maps values from high precision to low precision format.

\paragraph{Group Relative Policy Optimization}
Group Relative Policy Optimization (GRPO) 
is a value function-free variant of PPO for LLM post-training, 
where the objective for a training policy $\pi_\theta$ is derived directly from 
the relative rewards of multiple sampled responses generated by a rollout policy.
For a given prompt $q$ from the training dataset, 
the rollout policy $\mu_{\theta_\text{old}}$ 
samples a group of $G$ responses $\{o_1,\dots,o_G\}$, 
where $o_i \sim \mu_{\theta_\text{old}}(\cdot \mid q)$.
Each response $o_i$ receives a scalar reward based on task correctness 
or a verifiable objective.
To evaluate the current action, 
GRPO computes a group-relative advantage 
by normalizing rewards within the sampled group,
eliminating the need to fit a value function.
The GRPO objective is:
\begin{equation}
\label{eq:grpo}
\begin{aligned}
J(\theta)
= \mathbb{E}_{q, o_i \sim \mu_{\theta_\text{old}}(\cdot| q)}\!\Biggl[
\frac{1}{G}\sum_{i=1}^{G}\frac{1}{|o_i|}\sum_{t=1}^{|o_i|}\min\!\big(
R_{i,t}A_{i,t},\,
\operatorname{clip}(R_{i,t},1-\epsilon_{\text{low}},1+\epsilon_{\text{high}})A_{i,t}
\big)
\Biggr].
\end{aligned}
\end{equation}
where
$A_{i,t}$ is the token-level advantage derived from the rewards, $R_{i,t}=\frac{\pi_\theta(o_{i,t}\mid q,o_{i,<t})}
{\pi_{\theta_\text{old}}(o_{i,t}\mid q,o_{i,<t})}$ is the importance sampling ratio , $\pi_\theta$ and  $\pi_{\theta_{\text{old}}}$ are the target and reference policies respectively, $\epsilon_{\text{low}}$ and $\epsilon_{\text{high}}$ are the clipping parameters. Note that, the objective also contains a KL penalty term but is omitted here for simplicity. 
\section{Method}





\subsection{Problem Setting}
Following previous works~\cite{qerl,qurl,jetrl},
we consider the standard Markov decision process (MDP) setting 
for LLMs post-training, 
where generation is viewed as a sequential decision-making problem. 
At each timestep, 
the state consists of the prompt together 
with the previously generated tokens (prefix), 
the action is the next token, 
and the episode ends when an end-of-sequence (EOS) token is produced 
or a maximum response length is reached. 

Our focus is fully quantized RL post-training, 
where both rollout and training are performed using quantized policies. 
We study this setting in the GRPO-based post-training framework 
implemented in VeRL~\cite{sheng2024hybridflow},
using HiF4~\cite{luo2026hifloat4} and 
MXFP4~\cite{darvishrouhani2023microscaling} 
as the FP4 instantiations of $Q$.

\begin{wraptable}{r}{0.5\textwidth}
    \centering
    \footnotesize
    \setlength{\tabcolsep}{4pt}
    \begin{tabular}{llcc}
        \toprule
        & & \multicolumn{2}{c}{Accuracy (\%)} \\
        \cmidrule(lr){3-4}
        Train & Rollout & FP4 = HiF4 & FP4 = MXFP4 \\
        \midrule
        BF16 & BF16  & \multicolumn{2}{c}{86.96} \\
        \addlinespace[2pt]
        FP4  & BF16  & 86.88 & 86.20 \\
        FP4  & W4A16 & 85.22 & 81.57 \\
        FP4  & W16A4 & 84.30 & 77.71 \\
        FP4  & FP4   & 82.03 & 73.31 \\
        BF16 & FP4   & 78.01 & 50.57 \\
        \bottomrule
    \end{tabular}
    \caption{
    Accuracy on GSM8K test set for Qwen2.5-3B with GRPO under different training and rollout precisions. FP4 formats include HiF4 and MXFP4, with W4A16 and W16A4 denoting weight-only and activation-only quantization, respectively.
    }
    \label{tab:train_rollout_ablation}
    \vspace{-8pt}
\end{wraptable}

\subsection{The Cause of Performance Degradation in FP4 RL}
\label{sec:issue_fp4_rollout}

Table~\ref{tab:train_rollout_ablation} presents a comparison of training results for Qwen2.5-3B on GSM8K across different precision configurations for training and rollout. 
In the full BF16 baseline (Row 1), the model achieves $86.96\%$ accuracy. When both training and rollout are in HiF4 (Row \review{5}), the accuracy drops to $82.03\%$. This degradation is even more pronounced for MXFP4, where accuracy falls to $73.31\%$. These results raise a fundamental question for low-precision RL --- is the loss in performance driven by training quantization or rollout quantization?  
To identify the source of this degradation, we compare the performance of applying quantization to specific components. Our result indicates that
the gap between full BF16 RL and full FP4 RL
is primarily driven by rollout quantization.
Specifically, 
training in HiF4 or MXFP4 while maintaining a BF16 rollout (Row 2) yields accuracies of $86.88\%$ and $86.20\%$, respectively — matching the full BF16 baseline. In other words, by simply keeping rollout in high precision, the accuracy is recovered to nearly fully high precision level. 
Conversely, 
applying FP4 quantization to the rollout phase 
while keeping the training in BF16 (Row \review{6}) results in a catastrophic drop to $78.01\%$ for HiF4 and $50.57\%$ for MXFP4, 
performing significantly worse than even the fully quantized settings. \review{ We observe that the KL divergence between the inference and training policies throughout the RL training is substantially higher than all the other settings (see Appendix~\ref{app:kl-div}), indicating that policy mismatch is the primary mode of failure if only training is in BF16.}
Furthermore, by isolating the components of the rollout phase, 
we identify activation quantization as the primary bottleneck. 
While weight-only quantization (W4A16) in the rollout (Row 3) 
maintains a competitive accuracy of $85.22\%$ for HiF4 and $81.57\%$ for MXFP4, 
activation-only quantization (W16A4, Row \review{4}) leads to a sharper decline to $84.30\%$ for HiF4 and $77.71\%$ for MXFP4.
These findings reveal that, in low-precision RL,
the dominant source of model degradation
is the quantization error of activations during rollout, 
rather than the precision loss due to quantization during training.

\begin{figure}[tb]
    \centering
    \begin{minipage}[t]{0.48\textwidth}
        \centering
        \vspace{0pt} 
        \includegraphics[width=\textwidth]{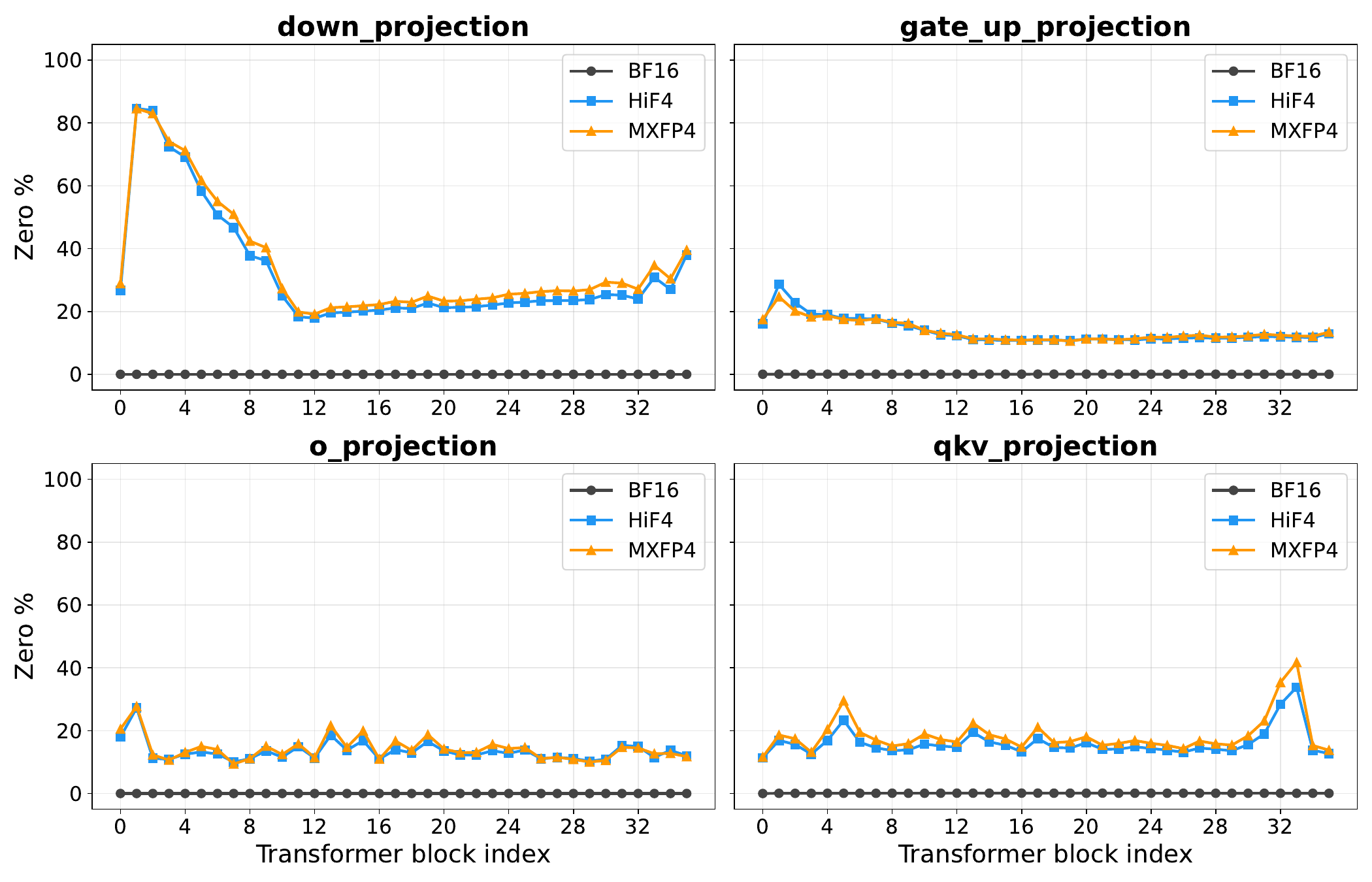}
        \caption{Percentage of zero values in activation tensors of each layer before quantization (BF16) and after quantization (HiF4, MXFP4) with GSM8K and Qwen2.5-3B.}
        \label{fig:activation_zero_percent}
    \end{minipage}
    \hfill
    \begin{minipage}[t]{0.48\textwidth}
        \centering
        \vspace{0pt} 
        \includegraphics[width=\textwidth]{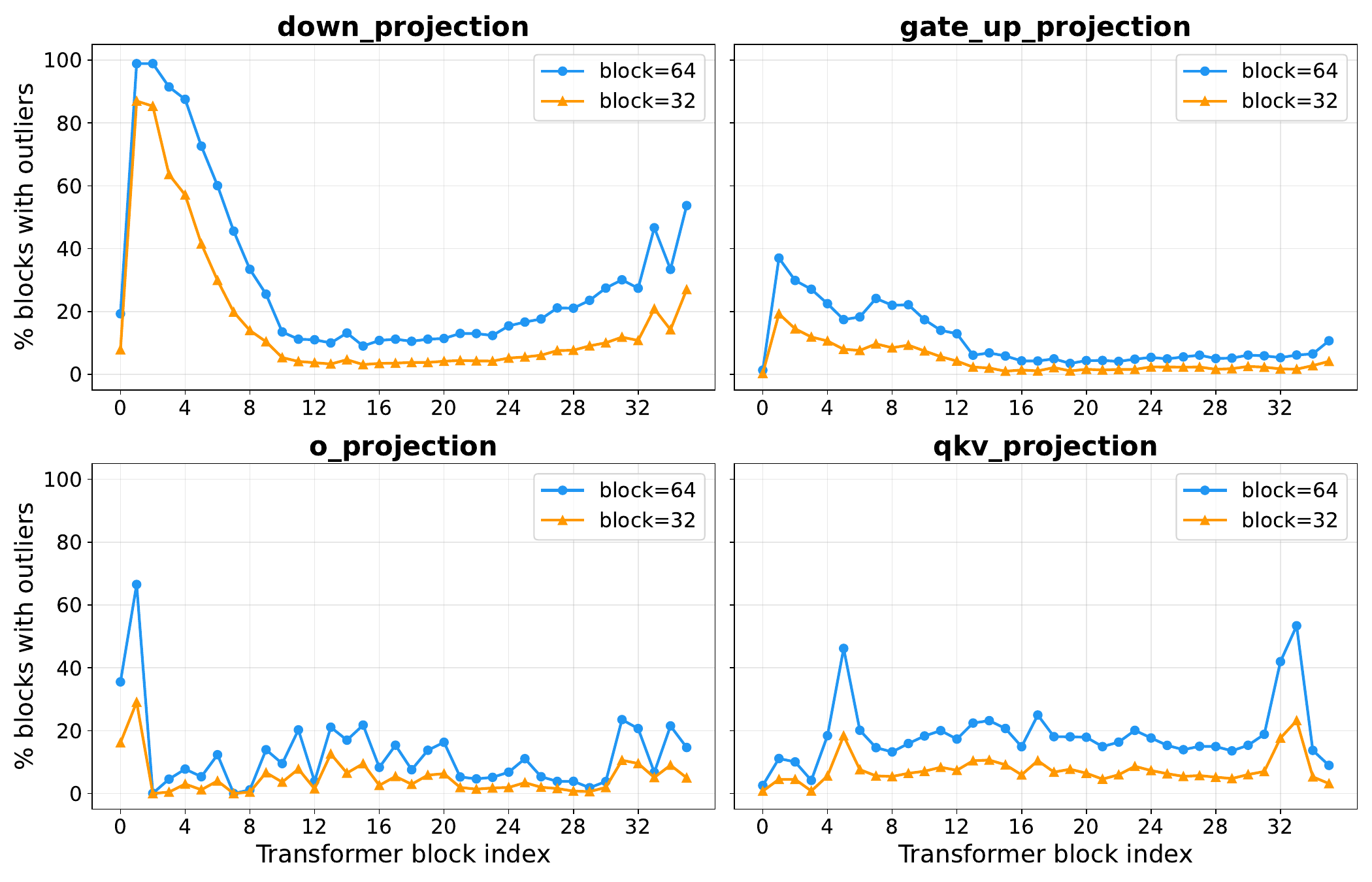}
        \caption{Percentage of blocks with outliers in activation tensors of each layer with GSM8K and Qwen2.5-3B for HiF4 and MXFP4 block size.}
        \label{fig:activation_outlier_blk_percent}
    \end{minipage}
    \vspace{-16pt}
\end{figure}

To further understand the issue with rollout activation, we analyze the distribution of the rollout activation tensors. Figure \ref{fig:activation_zero_percent} illustrates the percentage of zero values within activation tensors across layers. While the original BF16 tensors are dense, post-quantization tensors (HiF4, MXFP4) show a dramatic increase in sparsity. In specific layers, such as 
the down projection of transformer block 1 and 2, 
sparsity exceeds $80\%$, indicating a "zero collapse" where the majority of non-zero values are mapped to zero. This phenomenon is driven by the presence of outliers, as emphasized in \cite{wang2025optimizing} and \cite{zhang2025accurate}. As shown in Figure \ref{fig:activation_outlier_blk_percent}, the percentage of blocks containing outliers strongly correlates with the severity of zero collapse in the activation tensor. During quantization, since the block scale factor is governed by the large values in a block, outliers can lead to enormous scale factors that collapse the non-outliers to zeros. Consequently, the tensor's distribution collapses, resulting in significant information loss and precision degradation.

\begin{figure*}[t]
  \centering
  \begin{subfigure}[t]{0.35\linewidth}
    \centering
    \includegraphics[width=\linewidth]{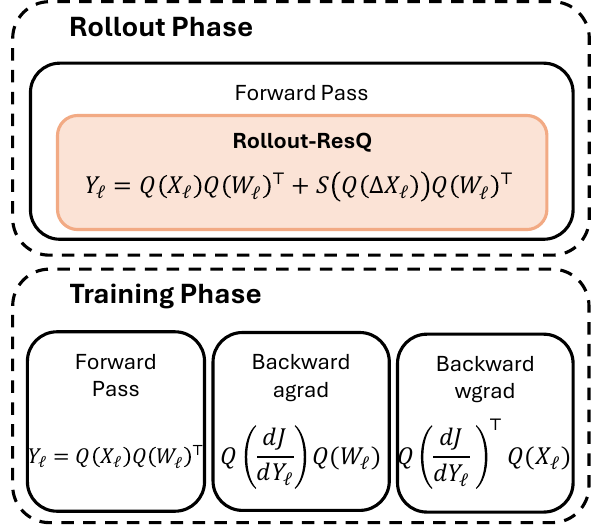}
    \caption{FP4 RL recipe with \ourmethod}
    \label{fig:ourmethod_framework_a}
  \end{subfigure}
  \begin{subfigure}[t]{0.64\linewidth}
    \centering
    \includegraphics[width=\linewidth]{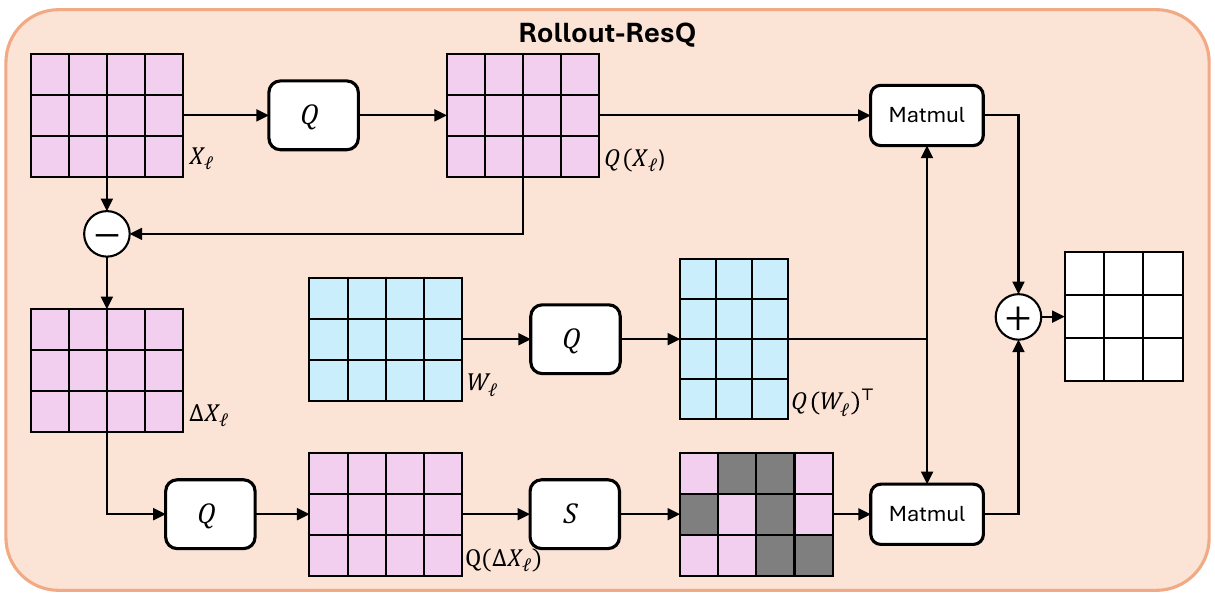}
    \caption{The computation of \ourmethod for rollout linear projections.}
    \label{fig:ourmethod_framework_b}
  \end{subfigure}
  \caption{\ourmethod overview.}
  \label{fig:ourmethod_framework}
\end{figure*}

\subsection{Rollout Residual Quantization (\ourmethod) for FP4 RL}
We present \ourmethod, an effective 
strategy 
for recovering precision loss 
caused by activation outliers during rollout in fully FP4 RL.
As shown in Figure~\ref{fig:ourmethod_framework_a},
\ourmethod is used in the rollout phase 
of our fully FP4 RL training recipe, 
augmenting the standard FP4 matrix multiplication 
in the linear projection layers of all transformer blocks with a residual correction term. 
Figure~\ref{fig:ourmethod_framework_b} illustrates 
this design in detail.
During rollout, 
for each linear layer $\ell$, 
the \ourmethod 
takes as inputs the activation tensor $X_\ell$ and 
weight tensor $W_\ell$ and 
computes an activation residual,
\begin{equation}
\Delta X_\ell = X_\ell - Q(X_\ell),
\end{equation}
where $Q$ is a quantization method. 
The residual captures the activation quantization error 
which can then be quantized to recover the precision loss through two FP4 GEMMs,
\begin{equation}
\label{eq:of}
\hat{Y_\ell}
=
Q(X_\ell)Q(W_\ell)^\top
+
Q(\Delta X_\ell)Q(W_\ell)^\top ,
\end{equation}
Note that, unlike~\cite{zhang2025accurate} which computes residuals at block level, this formulation operates on a tensor level, which can be implemented more efficiently with modern deep learning framework.
However, this approach still requires an extra dense matrix multiplication 
which can incur substantial computation cost. 
To maintain computation efficiency, 
\ourmethod utilizes a sparsity function $S$, 
which is applied to the quantized activation residual $Q(\Delta X_{\ell})$.
The resulting rollout projection is defined as:
\begin{equation}
\label{eq:resq}
\hat{Y_\ell}
=
Q(X_\ell)Q(W_\ell)^\top
+
S(Q(\Delta X_\ell))Q(W_\ell)^\top ,
\end{equation}
By integrating these components, 
Equation~\ref{eq:resq} preserves 
the efficiency of the standard FP4 projection 
while introducing a lightweight, sparse matrix multiplication,
which
is intended to reduce computation and 
can provide faster execution, 
especially when its granularity matches 
hardware-supported 
sparse kernels~\cite{MLSYS2023_5a54f793,10.1145/3005348,NEURIPS2021_6e8404c3,DBLP:conf/hpca/LiuWZM22}.
We instantiate $S$ using different sparsity functions:
(1) structured channel-wise sparsity, denoted by $S_{\alpha\%}$,
which retains only the top $\alpha\%$ of channels 
with the largest L2-norm;
(2) semi-structured sparsity, denoted by $S_{M:N}$,
which retains only the $M$ largest magnitude elements
within every contiguous block of $N$ elements; and
(3) block sparsity, denoted by $S_{\alpha\%}^{M{\times}N}$,
which retains the top $\alpha\%$ of blocks of size $M{\times}N$ 
with largest L2-norm.
These sparsity functions are motivated by prior work 
showing the practical benefits of 
channel-wise, 
semi-structured, 
and block sparsity 
for efficient 
matrix multiplication~\cite{cho2023pdp,10.1145/3005348,NEURIPS2021_6e8404c3,DBLP:conf/hpca/LiuWZM22,mishra2021acceleratingsparsedeepneural,haziza2025accelerating,9857911,pmlr-v234-ma24a}.
In our experiments, 
we choose $\alpha=50$, a commonly used pruning level 
in prior work~\cite{cho2023pdp}.
We choose $S_{2:4}$ since it is supported by modern sparse kernels and 
offers a trade-off between 
acceleration and accuracy~\cite{10.5555/3692070.3692857,haziza2025accelerating}\review{, and has been shown to be able to achieve a speedup of 1.3-1.5x~\cite{Meghana2026sparsity, haziza2025accelerating}}.
Finally, for block sparsity we choose $S^{32\times 64}_{50\%}$ where half of the $32\times 64$ blocks are kept.
In all three cases, we sparsify $Q(\Delta X_\ell)$ on a $50\%$ level.
We denote the channel-wise, 
semi-structured, and 
block-sparse variants of \ourmethod 
as \ourmethodc, \ourmethods, and \ourmethodb, respectively.
Note that setting $S$ to the identity function,
recovers Equation~\ref{eq:of};
we refer to this method as \ourmethodd.

\setlength\intextsep{0pt}
\begin{wrapfigure}{r}{0.5\textwidth} 
    \centering
    \includegraphics[width=0.5\textwidth]{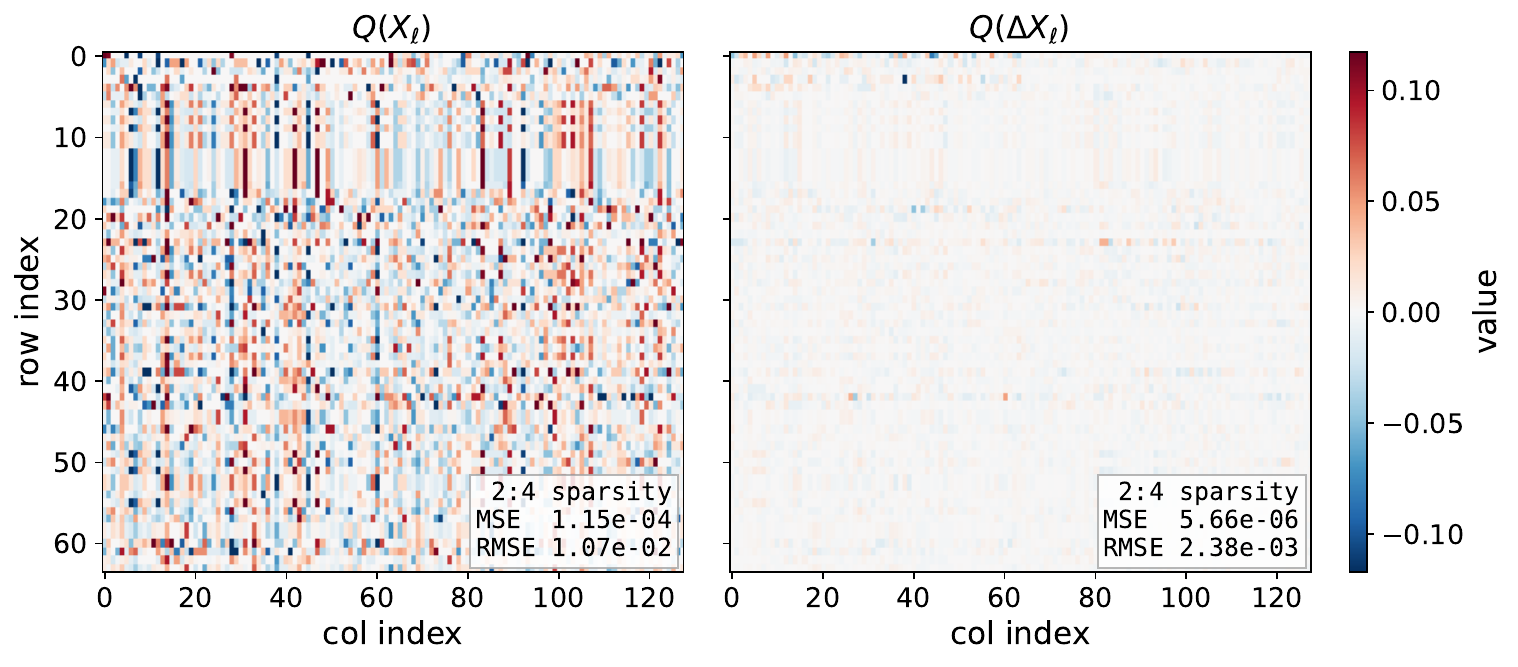}
    \caption{Heatmap of quantized activation and residual during rollout. The MSE and RMSE before and after performing 2:4 sparsity.}
    \label{fig:heatmap_activation_residual}
    \vspace{-11pt}
\end{wrapfigure}

The choice to sparsify the residual rather than the primary activation is motivated by their robustness to sparsification.
We measure MSE and RMSE of $Q(X_{\ell})$ and $Q(\Delta X_{\ell})$
with their corresponding $S_{2:4}$ sparsified versions. 
As illustrated in Figure~\ref{fig:heatmap_activation_residual}, our empirical measurements confirm that $Q(\Delta X_{\ell})$ is more robust to $S_{2:4}$, incurring smaller error, making it an ideal candidate for hardware-friendly acceleration.

\section{Experiments}

\begin{wraptable}{r}{0.6\textwidth}
\centering
\footnotesize
\setlength{\tabcolsep}{5pt}
\begin{tabular}{llll}
\toprule
\multirow{2}{*}{Format} & \multirow{2}{*}{Recovery Method} & \multicolumn{2}{c}{Accuracy} \\
\cmidrule(lr){3-4}
& & GSM8K-test & Math500 \\
\midrule
BF16 & --
& $86.96$ & $61.2$ \\
\midrule
\multirow{10}{*}{HiF4}
& --
& $82.03$ {\scriptsize ($-4.93$)} & $44.6$ {\scriptsize ($-16.6$)} \\
& SmoothQuant
& $78.54$ {\scriptsize ($-8.42$)} & $39.0$ {\scriptsize ($-22.2$)} \\
& RHT
& $82.87$ {\scriptsize ($-4.09$)} & $41.0$ {\scriptsize ($-20.2$)} \\
& OCC
& $84.23$ {\scriptsize ($-2.73$)} & $51.8$ {\scriptsize ($-9.4$)} \\
& \ourmethodd
& $85.52$ {\scriptsize ($-1.44$)} & $53.0$ {\scriptsize ($-8.2$)} \\
& \ourmethodc
& $84.61$ {\scriptsize ($-2.35$)} & $52.6$ {\scriptsize $(-8.6)$} \\
& \ourmethodb
& $84.61$ {\scriptsize ($-2.35$)} & $\boldsymbol{\textbf{$55.8$ \scriptsize ($-5.4$)}}$ \\
& \ourmethods
& \textbf{$\boldsymbol{85.90}$ {\scriptsize $\boldsymbol{(-1.06)}$}} & $52.6$ {\scriptsize ($-8.6$)} \\
\midrule
\multirow{10}{*}{MXFP4}
& --
& $73.31$ {\scriptsize ($-13.65$)} & $27.8$ {\scriptsize ($-33.4$)} \\
& SmoothQuant
& $64.22$ {\scriptsize ($-22.74$)} & $23.6$ {\scriptsize ($-37.6$)} \\
& RHT
& $78.01$ {\scriptsize ($-8.95$)} & $39.6$ {\scriptsize ($-21.6$)} \\
& OCC
& $78.70$ {\scriptsize ($-8.26$)} & $42.8$ {\scriptsize ($-18.4$)} \\
& \ourmethodd
    & $\boldsymbol{81.65}$ {\scriptsize $\boldsymbol{(-5.31)}$} & ${\boldsymbol{\textbf{$48.2$ {\scriptsize ($-13.0$)}}}}$ \\
& \ourmethodc
& $80.82$ {\scriptsize ($-6.14$)} & $42.8$ {\scriptsize ($-18.4$)} \\
& \ourmethodb
& $80.36$ {\scriptsize ($-6.60$)} & $45.6$ {\scriptsize ($-15.6$)} \\
& \ourmethods
& $\boldsymbol{81.65}$ {\scriptsize $\boldsymbol{(-5.31)}$} & $46.4$ {\scriptsize ($-14.8$)} \\
\bottomrule
\end{tabular}
\caption{Results for Qwen2.5-3B trained on GSM8K with GRPO. Values in parentheses show gaps to BF16.}
\label{tab:results_perf_3b}
\end{wraptable}
\subsection{Experimental Setup}
We conduct all experiments using the training framework via VeRL~\cite{sheng2024hybridflow},
a hybrid-engine architecture that enables different execution engines to be combined for improved hardware efficiency.
In our experiments, 
we use vLLM for the rollout policy and 
FSDP for the training policy,
with GRPO as the RL backbone.
We study two FP4 quantization format, HiF4 and MXFP4, 
applying quantization in the linear layers of all transformer blocks. 
We train Qwen2.5-3B~\cite{qwen2025qwen25technicalreport} on GSM8K~\cite{cobbe2021gsm8k}, 
and Qwen2.5-Math-7B~\cite{yang2024qwen25mathtechnicalreportmathematical} on DAPO-Math-17K~\cite{yu2026dapo}
which is a more challenging task.
See Appendix~\ref{app:training_details} for additional details.

For evaluation, 
we use several widely used mathematical reasoning benchmarks.
For Qwen2.5-3B, 
we report results on held-out GSM8K-test samples 
and on Math-500~\cite{hendrycks2021measuring}.
For Qwen2.5-Math-7B, 
we report results on AIME-2024~\cite{jia2024aime2024}, 
AIME-2025~\cite{aime2025_hf}, 
AMC-2023~\cite{10.5555/3600270.3602281}, 
and Math-500~\cite{hendrycks2021measuring}.
We report Mean@1 for Qwen2.5-3B and 
Mean@32 for Qwen2.5-Math-7B, 
where Mean@1 denotes accuracy under greedy decoding and 
Mean@32 denotes average accuracy over 32 sampled responses.
We compare all the variants of \ourmethod against several baselines:
(1) BF16, which uses full-precision BF16 for both rollout and training;
(2) HiF4 and MXFP4, which quantizes both rollout and training using the corresponding quantization method without any rollout recovery method;
(3) HiF4 and MXFP4 with SmoothQuant~\cite{xiao2023smoothquant}, RHT~\cite{tseng2025trainingllmsmxfp4} and OCC~\cite{wang2025optimizing} applied to rollout activations. In SmoothQuant experiments, the smoothing scales are calibrated at the start of each RL iteration.

\subsection{GSM8K Training Results}
\begin{figure*}[t]
  \centering
  \begin{subfigure}[t]{0.325\linewidth}
    \centering
    \includegraphics[width=\linewidth]{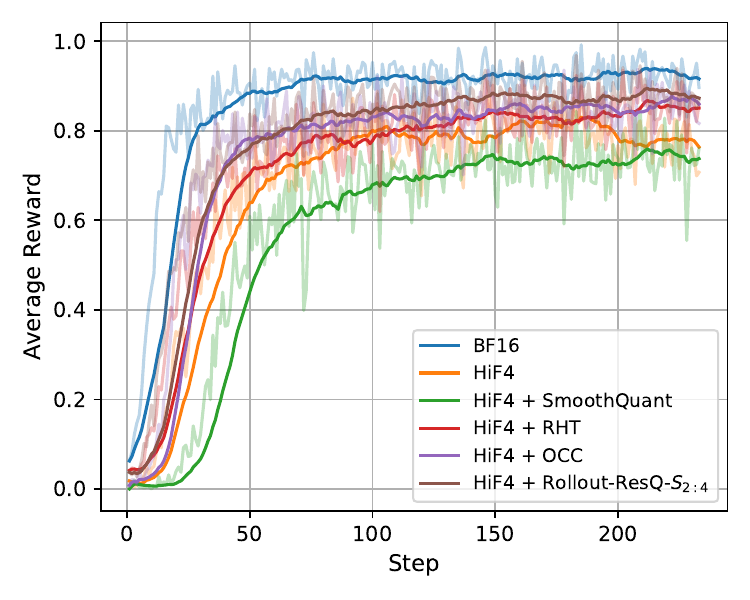}
    \caption{Qwen2.5-3B under HiF4}
    \label{fig:train_row_a}
  \end{subfigure}
  \begin{subfigure}[t]{0.325\linewidth}
    \centering
    \includegraphics[width=\linewidth]{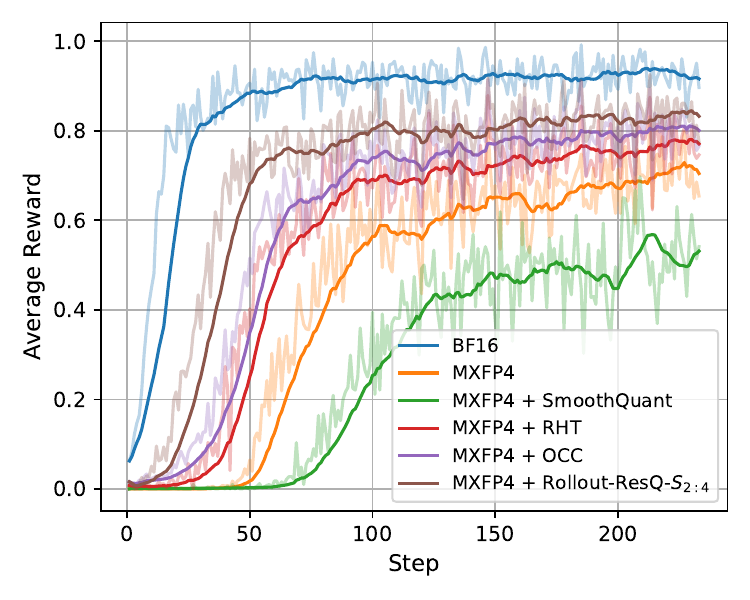}
    \caption{Qwen2.5-3B under MXFP4}
    \label{fig:train_row_b}
  \end{subfigure}
  \begin{subfigure}[t]{0.325\linewidth}
    \centering
    \includegraphics[width=\linewidth]{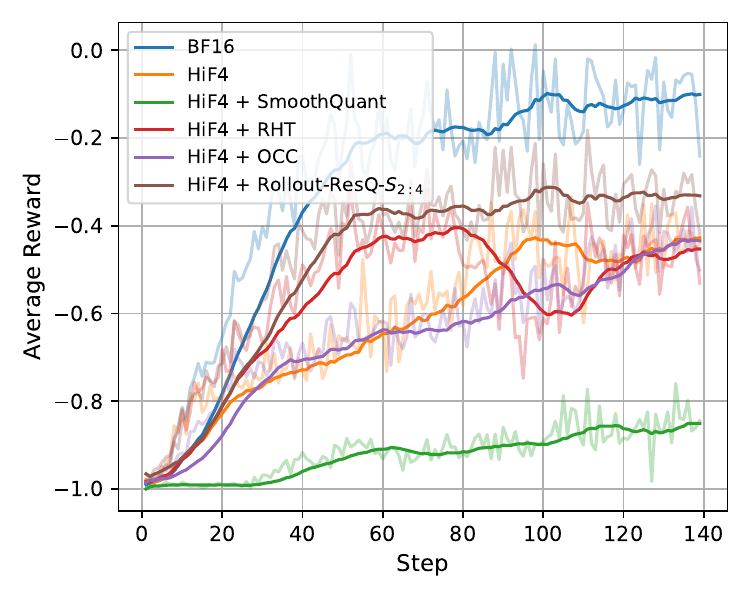}
    \caption{Qwen2.5-Math-7B under HiF4}
    \label{fig:train_row_c}
  \end{subfigure}
  \caption{Training performance (average reward) for Qwen2.5-3B on GSM8K under HiF4 and MXFP4, and for Qwen2.5-Math-7B on DAPO-Math-17K under HiF4.}
  \label{fig:train_result_row}
\end{figure*}

Table~\ref{tab:results_perf_3b} reports  
Mean@1 accuracies of 
training 3B model on GSM8K. 
Naively applying HiF4 and MXFP4 to low-precision RL training degrades performance by 4.93\% and 13.65\% on GSM8K-test, and 16.6\% and 33.4\% on MATH500, respectively, relative to BF16. 
Among the baselines, 
OCC~\cite{wang2025optimizing} performs better, 
but still leaves gaps of $2.73\%$ and $8.26\%$ on GSM8K-test,
and $9.4\%$ and $18.4\%$ on Math500, 
under HiF4 and MXFP4, respectively.
By contrast, 
\ourmethods reduces these gaps to $1.06\%$ and $5.31\%$ on GSM8K-test 
for HiF4 and MXFP4, respectively.
On Math500, under HiF4, \ourmethodb achieves the highest accuracy (55.8\%), followed by other \ourmethod variants.
Under MXFP4, \ourmethodd leads (48.2\%), followed by \ourmethods{}(46.4\%) and \ourmethodc{}(45.6\%). Across both formats, all \ourmethod variants outperform all baselines and close the gap to BF16 significantly.
\review{ Overall, these results show that \ourmethod substantially outperforms the naive-FP4 baseline on both formats and closes the accuracy gap to BF16 more effectively than other baseline methods.
}
The consistent advantage of \ourmethod over all baselines under both HiF4 and MXFP4 confirms that the sparse residual correction is format-agnostic. In all settings, 
\review{ HiF4 establishes a significantly higher baseline and recovery ceiling than MXFP4 due to its hierarchical scaling, }
revealing that HiF4 is the more robust format for end-to-end FP4 RL post-training. \review{ And Rollout-ResQ acts as the crucial mechanism for closing the remaining gap to BF16 across both formats -- providing substantial gains even on the weaker format (MXFP4).}
Figure~\ref{fig:train_row_a} and~\ref{fig:train_row_b} present 
the training average reward for the 3B model on HiF4 and MXFP4, respectively.
The figure reports the best-performing variant of \ourmethod, i.e., \ourmethods, 
during the training.
We make the following observations:
(1) 
Under both HiF4 and MXFP4, SmoothQuant exhibits slow and suboptimal convergence, which we attribute to its reliance on accurate calibration scales — a requirement that is hard to satisfy under RL's non-stationary rollout distribution.
(2) RHT and OCC consistently outperform SmoothQuant, 
including the early stage of training. 
(3) Under both HiF4 and MXFP4, 
\ourmethods consistently outperforms the baselines across training and achieves substantially faster convergence;
(4) HiF4 appears to be more compatible with both \ourmethods and the baselines, 
as MXFP4 leads to less stable training dynamics across methods.


\review{ We also performed multi-seed experiments for BF16, HiF4, OCC, and \ourmethods to confirm the stability of relative effectiveness of these methods. 
Table~\ref{tab:3b-seed-mean} reports the means and sample standard deviations of GSM8K-test and Math500 accuracy for 3-seed runs of BF16, HiF4, OCC, and \ourmethods for Qwen2.5-3B with GSM8K training. 
The results are consistent with our 1-seed experiments, where the HiF4 baseline exhibits the most degradation (-3.59\% on GSM8K-test, -13.87\% on Math500), and \ourmethods (-1.39\% on GSM8K-test, -6.94\% on Math500) improves over OCC (-2.13\% on GSM8K-test, -8.80\% on Math500), with a clear gain on Math500 and a modest gain on GSM8K-test.
}


\begin{table}[htbp]
  \centering
  \small 
  
  \label{tab:3b-seed-mean}
  \vspace{6pt}
  \renewcommand{\arraystretch}{1.2}
  \begin{tabular}{l|ll|ll}
    \hline
    & \multicolumn{2}{c|}{GSM8K-test} & \multicolumn{2}{c}{Math500} \\
    \cline{2-5}
    Qwen2.5-3B & Mean & Std. & Mean & Std. \\
    \hline
    BF16 & 86.56 & 0.42 & 60.47 & 1.63 \\
    HiF4 & 82.97 (-3.59) & 0.88 & 46.60 (-13.87) & 2.11 \\
    OCC & 84.43 (-2.13) & 0.41 & 51.67 (-8.80) & 0.42 \\
    \ourmethods & 85.17 (-1.39) & 0.64 & 53.53 (-6.94) & 0.81 \\
    \hline
  \end{tabular}
  \vspace{5pt}
  \caption{Results of Qwen2.5-3B trained on GSM8K with GRPO over 3 seeds for BF16, HiF4, OCC and \ourmethods in the HiF4 context. The mean and sample standard deviation of GSM8K-test and Math500 are reported. Values in parentheses show gaps to BF16.}
  \label{tab:3b-seed-mean}
\end{table}

\subsection{DAPO-Math-17K Training Results}


We focus on HiF4 for the 7B model, as the 3B results in Section 5.2 indicate that HiF4 is a more reliable format for FP4 RL.
Table~\ref{tab:results_perf_7bmath} 
reports 
Mean@32 accuracies of training Math-7B model on DAPO-Math-17K.
The table shows that 
naive HiF4 incurs drops of $11.14\%$, $4.79\%$, $9.37\%$ and $15.19\%$,
relative to full-precision BF16.
Consistent with the previous section,
SmoothQuant performs poorly in this setting.
In contrast, \ourmethods narrows the gaps to BF16 to
$6.56\%$, $3.54\%$, $8.75\%$, and $5.64\%$, respectively,
with \ourmethodd performs slightly better on AIME25 and Math500. 
Overall, \ourmethods offers the most consistent improvement across all benchmarks.
These results highlight both the effectiveness of \ourmethods 
and its generalization to diverse math benchmarks.
Figure~\ref{fig:train_row_c} shows the training average reward of the Math-7B model.
We make the following observations:
(1) Most baseline methods exhibit either unstable behavior or slow convergence. In contrast, \ourmethods remains relatively stable throughout training and achieves higher performance;
(2) Although RHT is competitive with \ourmethods early in training, its reward begins to deteriorate later on, indicating less stable training dynamics;
(3) OCC shows a similar training progress to naive HiF4, but this does not translate to downstream performance --- exhibiting a significantly drop-off from HiF4 in all four benchmarks in Table~\ref{tab:results_perf_7bmath}. 
(4) \ourmethods reaches a higher reward level earlier than baselines and maintains it more consistently, exhibiting better convergence speed and stability.


\begin{table*}[htbp]
\vspace{8pt}
\centering
\footnotesize
\setlength{\tabcolsep}{5pt}
\begin{tabular}{llllll}
\toprule
\multirow{2}{*}{Precision} & \multirow{2}{*}{Recovery Method} & \multicolumn{4}{c}{Accuracy (Mean@32)} \\
\cmidrule(lr){3-6}
& & AIME24 & AIME25 & AMC23 & Math500 \\
\midrule
BF16 & --
& $27.29$ & $12.08$ & $64.53$ & $74.59$ \\
\midrule
\multirow{8}{*}{HiF4}
& --
& $16.15$ {\scriptsize $(-11.14)$} & $7.29$ {\scriptsize $(-4.79)$} & $55.16$ {\scriptsize $(-9.37)$} & $59.40$ {\scriptsize $(-15.19)$} \\
& SmoothQuant
& $0.70$ {\scriptsize $(-26.59)$} & $0.10$ {\scriptsize $(-11.98)$} & $23.36$ {\scriptsize $(-41.17)$} & $53.51$ {\scriptsize $(-21.08)$} \\
& RHT
& $13.43$ {\scriptsize $(-13.86)$} & $8.64$ {\scriptsize $(-3.44)$} & $53.75$ {\scriptsize $(-10.78)$} & $64.67$ {\scriptsize $(-9.92)$} \\
& OCC
& $6.77$ {\scriptsize $(-20.52)$} & $4.90$ {\scriptsize $(-7.18)$}& $41.33$ {\scriptsize $(-23.20)$}& $43.59$ {\scriptsize $(-31.00)$}\\
& \ourmethodd
& $14.48$ {\scriptsize $(-12.81)$} & $\boldsymbol{8.96}$ {\scriptsize $\boldsymbol{(-3.12)}$} & $55.62$ {\scriptsize $(-8.91)$} & $\boldsymbol{69.02}$ {\scriptsize $\boldsymbol{(-5.57)}$}\\
& \ourmethodc
& $13.44$ {\scriptsize $(-14.46)$} & $7.60$ {\scriptsize $(-4.48)$} & $43.51$ {\scriptsize $(-21.01)$} & $64.30$ {\scriptsize $(-10.29)$}\\
& \ourmethodb
& $14.06$ {\scriptsize $(-13.84)$} & $8.33$ {\scriptsize $(-3.75)$} & $53.28$ {\scriptsize $(-11.25)$} & $47.89$ {\scriptsize $(-26.70)$}\\
& \ourmethods
& $\boldsymbol{20.73}$ {\scriptsize $\boldsymbol{(-6.56)}$} & $8.54$ {\scriptsize $(-3.54)$} & $\boldsymbol{55.78}$ {\scriptsize $\boldsymbol{(-8.75)}$} & $68.95$ {\scriptsize $(-5.64)$} \\
\bottomrule
\end{tabular}
\caption{Results for Qwen2.5-Math-7B trained on DAPO-Math-17K with GRPO. Parentheses show gaps to the BF16 baseline.}
\label{tab:results_perf_7bmath}
\end{table*}


\subsection{Overhead Analysis}
We analyze the theoretical computational overhead 
introduced by \ourmethod, specifically \ourmethods.
For a linear layer with input activation 
$X \in \mathbb{R}^{m \times d}$ and 
weight matrix $W \in \mathbb{R}^{n \times d}$, 
a dense FP16/BF16 GEMM for computing $Y = XW^\top$ has a FLOP cost of $2mdn$.
In \ourmethods, the rollout projection 
is computed as the sum of a dense GEMM and a sparse residual GEMM.
Under the $S_{2:4}$ sparsity function, 
only half of the residual activation elements are retained, 
so the number of nonzero entries is $\tfrac{1}{2}md$. 
Multiplying this sparse residual by $W^\top$ costs $mdn$ FLOPs.
Assuming FP4 execution provides 
a $4\times$ throughput advantage 
over FP16 on supported hardware~\cite{pareto,pact,ultra4bit,dpquant},
the effective costs of the dense and sparse GEMMs 
become $\tfrac{1}{2}mdn$ and $\tfrac{1}{4}mdn$, respectively.
Hence, the total effective compute cost of \ourmethodd is $\tfrac{1}{2}mdn + \tfrac{1}{2}mdn = mdn$ and \ourmethods is
$\tfrac{1}{2}mdn + \tfrac{1}{4}mdn = \tfrac{3}{4}mdn$.
Relative to FP16, while \ourmethodd achieves a theoretical speedup of $\tfrac{2mdn}{mdn} = 2\times$, \ourmethods can achieve up to $\tfrac{2mdn}{3mdn/4} \approx 2.67\times$ speedup.
Overall, \ourmethods and its variants introduce minimal overhead, which is critical for preserving FP4 efficiency.
\review{ We provide more analysis on the memory overhead of \ourmethods in Appendix~\ref{app:memory}}.



\section{Limitations and Future Directions} 
Due to the lack of native FP4 hardware support, all experiments rely on FP4 simulations\footnote{The simulated quantization code is publicly available at 
\url{https://github.com/global-computing-consortium/HiFloat4}.
} 
where simulation overhead dominates training time, preventing direct measurement of real speedup and efficiency gains. Additionally, the prolonged runtime has prevented us from extending experiments to extremely large models or datasets, which remain important directions for future work. 
Furthermore, this work focuses exclusively on 50\% sparsity across all sparsity patterns, with the $S_{2:4}$ variant driven by modern hardware support for semi-structured sparsity; exploring varying sparsity levels to better characterize the accuracy-efficiency trade-off remains an open direction.



\section{Conclusion}

This work presents the first end-to-end FP4 RL post-training for LLMs, with both rollout and training policies operating at 4-bit precision. We show that the central obstacle is not the training pass but the rollout: activation outliers lead to unwanted underflow in FP4 setting, degrading rollout quality and learning outcomes. 
To recover accuracy without significantly compromising compute efficiency, we proposed \ourmethod{} --- a sparse residual correction appended to the FP4 rollout matrix multiplication that neutralizes outlier-driven error while preserving all GEMMs in low precision. 
\review{\ourmethod{}} substantially closes the gap on Qwen2.5-3B and Qwen2.5-Math-7B across benchmarks, outperforming all outlier-mitigation baselines. We further show empirically that HiF4 is the recommended format for end-to-end FP4 RL, achieving a substantially tighter gap to full precision than MXFP4. These results demonstrate that robust fully FP4 RL post-training is possible when the dominant source of error --- rollout activation quantization --- is sufficiently addressed, and that format choice plays a key role in determining the recoverable accuracy.


\newpage

\bibliographystyle{plainnat}
\bibliography{citations.bib}

\appendix
\section{Training Details}
\label{app:training_details}
\paragraph{GRPO Training Configurations} 
Table \ref{tab:grpo_configs} summarizes the hyperparameters of GRPO training for Qwen2.5-3B and Qwen2.5-Math-7B experiments.

\begin{table}[ht]
    \centering
    \small 
    \caption{Hyperparameters for GRPO Training}
    \label{tab:grpo_configs}
    \begin{tabular}{@{}lcc@{}} 
        \toprule
        \textbf{Hyperparameter} & \textbf{Qwen2.5-3B} & \textbf{Qwen2.5-Math-7B} \\ 
        \midrule
        Policy learning rate           & $1 \times 10^{-6}$  & $1 \times 10^{-6}$       \\
        Training batch size              & 32                 & 128                      \\
        Group Size ($G$)        & 16                   & 16                       \\
        Max Prompt Length       & 512                 & 2048                     \\
        Max Response Length     & 1024                & 4096                     \\
        Optimizer               & AdamW               & AdamW                    \\
        Clip range $\epsilon_{\text{low}}$, $\epsilon_{\text{high}}$ & 0.2, 0.2 & 0.2, 0.28 \\
        \bottomrule
    \end{tabular}
    \vspace{4pt}
\end{table}

In all experiments, we apply a token-level truncated importance sampling threshold~\cite{yao2025on} of 5.0.


\paragraph{Compute Resources} All experiments were conducted on an internal cluster. The Qwen2.5-3B experiments consumed around 100 GB of peak memory and took approximately 10 hours to run. The Qwen2.5-Math-7B experiments consumed around 370 GB of peak memory and took approximately 20 hours to run. For OCC experiments, the runtime was notably longer due to its global top-percentile computation over the entire tensor.

\section{Memory Overhead of \ourmethods}\label{app:memory}

Per linear layer, let $X\in \mathbb{R}^{M\times K}, W\in \mathbb{R}^{K\times N}$ with $M=1$ per sequence. Rollout-ResQ-$S_{2:4}$ computes the quantized activation $Q(X)$, the sparse quantized residual $S(Q(\Delta X))$, the quantized weight $Q(W)$ with the sparse branch (density $d=0.5$) stored as a bitmask plus values.

We write $B_m^c$ for the per-layer memory traffic (in bytes) of component $c\in \{W, X\}$ (weight, activation) under method $m\in \{BF16, ours\}$ ($ours$ refers to Rollout-ResQ-$S_{2:4}$), and $B_m=B_m^W+B_m^X $ for the total.

\textbf {Weight}. $B_{BF16}^W=2KN$, $B_{ours}^W=\frac{1}{2}KN$, giving a $4\times$ reduction ($\frac{1}{2}KN/2KN=\frac{1}{4}$).

\textbf {Activation}. $B_{BF16}^X=2MK$, $B_{ours}^X=\underbrace{\frac{4}{8}MK}_{Q(X)}+\underbrace{\frac{MK}{8}}_{\text{mask}}+\underbrace{\frac{4}{8}dMK}_{S(Q(\Delta X))}=MK(\frac{5}{8}+\frac{1}{4})=\frac{7}{8}MK$, giving a $2.29\times $ reduction ($\frac{\frac{7}{8}MK}{2MK}=\frac{7}{16}$).

\textbf{Total ratio and decode regime}. Summing weight and activation traffic, the total memory relative to BF16 is

$$
\frac{B_{ours}}{B_{BF16}}=\frac{\frac{7}{8}MK +\frac{1}{2}KN}{2MK+2KN}=\frac{\frac{7}{8}M+\frac{1}{2}N}{2M+2N}.
$$

In the decode phase when $M=1$, this gives

$$
\frac{B_{ours}}{B_{BF16}}=\frac{\frac{7}{8}+\frac{N}{2}}{2(N+1)}=\frac{7}{16(N+1)}+\frac{1}{4}\cdot \frac{1}{1+1/N}.
$$

As $N$ grows, the first term vanishes and the second approaches $\frac{1}{4}$, so $\frac{B_{ours}}{B_{BF16}}\rightarrow  \frac{1}{4}$: a $4\times $ reduction, matching the FP4 baseline.

Forming $\Delta X=X-Q(X)$ needs the high-precision $X$ only transiently: it is already present as the incoming activation. A single pass consumes it and emits $Q(X)$ and $S(Q(\Delta X))$, after which $X$ is freed. No separate high-precision copy is stored, so the persistent activation footprint remains the $\frac{7}{8}MK$ above.

\section{Ablation Studies on Sparsity Patterns and Sparsity Levels}
We conducted an ablation study of different sparsity levels ($25\%, 50\%, 75\%$) for the three sparsity patterns \ourmethodb, \ourmethodc, \ourmethods.
\begin{table}[ht]
    \centering
    \small 
    \caption{GSM8K-test scores on different sparsity levels.}
    \label{tab:grpo_configs}
    \begin{tabular}{@{}lccc@{}} 
        \toprule
        \textbf{Density level} & \ourmethodb & \ourmethodc & \ourmethods \\ 
        \midrule
        $25\%$ & 84.53 & 84.31 & 84.31\\
        $50\%$ & 84.61 & 84.61 & 85.90\\
        $75\%$ & 84.99 & 84.84 & 85.90\\
        \bottomrule
    \end{tabular}\label{tab:sparse-level}
    \vspace{4pt}
\end{table}

Table~\ref{tab:sparse-level} shows that Rollout-ResQ is relatively robust to the choice of sparsity level, where the values are GSM8K-test scores after training. While increasing density from $25\%$ to $75\%$ leads to modest accuracy improvements, the gains beyond $50\%$ are small. The $50\%$ setting provides a favorable trade-off, achieving a significant inference speedup while matching the highest accuracy observed at $75\%$ (for the S2:4 configuration). Therefore, we use $50\%$ as the default setting in our experiments. Lower density levels can lead to reduced accuracy or unpredictable results. Based on our empirical observations, $50\%$ density was the most stable setting, providing significant reductions in memory usage and improved inference speed while preserving accuracy comparable to that of higher-density or dense matrices. 

\section{Comparison with High Precision Gradient via LoRA (QeRL)}
Tables~\ref{tab:qerl_3b} and~\ref{tab:qerl_7b} show that, on the 3B model, QeRL improves over the HiF4 baseline on both benchmarks and achieves the best MATH500 score overall. This pattern does not hold at 7B, however: QeRL not only trails \ourmethods across all four benchmarks but also falls below the HiF4 baseline on three of them---AIME24, AIME25, and AMC23---improving on HiF4 only on MATH500.

\begin{table}[ht]
    \centering
    \small
    \caption{Accuracy (\%) on Qwen2.5-3B.}
    \label{tab:qerl_3b}
    \begin{tabular}{@{}ccc@{}} 
        \toprule
        \textbf{Method} & GSM8K-test & MATH500 \\
        \midrule
        HiF4        & 82.03          & 44.60 \\
        QeRL        & 83.92          & \textbf{52.80} \\
        \ourmethods & \textbf{85.90} & 52.60 \\
        \bottomrule
    \end{tabular}
    \vspace{4pt}
\end{table}

\begin{table}[ht]
    \centering
    \small
    \caption{Accuracy (\%) on Qwen2.5-7B.}
    \label{tab:qerl_7b}
    \begin{tabular}{@{}ccccc@{}} 
        \toprule
        \textbf{Method} & AIME24 & AIME25 & AMC23 & MATH500 \\
        \midrule
        HiF4        & 16.15          & 7.29          & 55.16          & 59.40 \\
        QeRL        & 12.29          & 4.69          & 47.19          & 67.65 \\
        \ourmethods & \textbf{20.73} & \textbf{8.54} & \textbf{55.78} & \textbf{68.95} \\
        \bottomrule
    \end{tabular}
    \vspace{4pt}
\end{table}

\section{KL Divergence between Inference and Training Policies}
\label{app:kl-div}
\begin{figure}[htbp]
    \centering
    \begin{subfigure}[b]{0.48\textwidth}
        \centering
        \includegraphics[width=\textwidth]{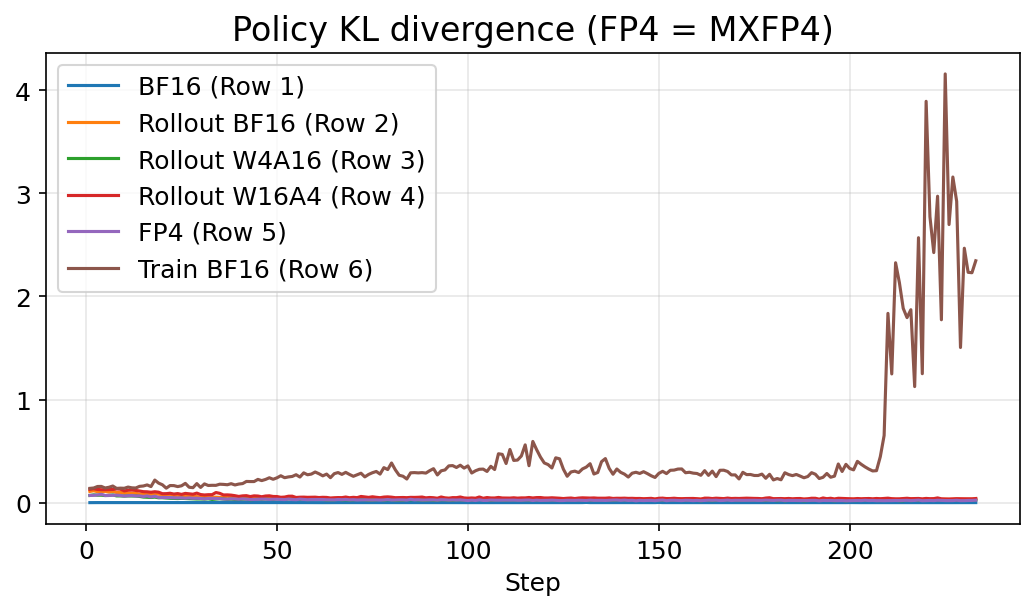}
        \caption{HiF4}
        
    \end{subfigure}
    \hfill 
    \begin{subfigure}[b]{0.48\textwidth}
        \centering
        \includegraphics[width=\textwidth]{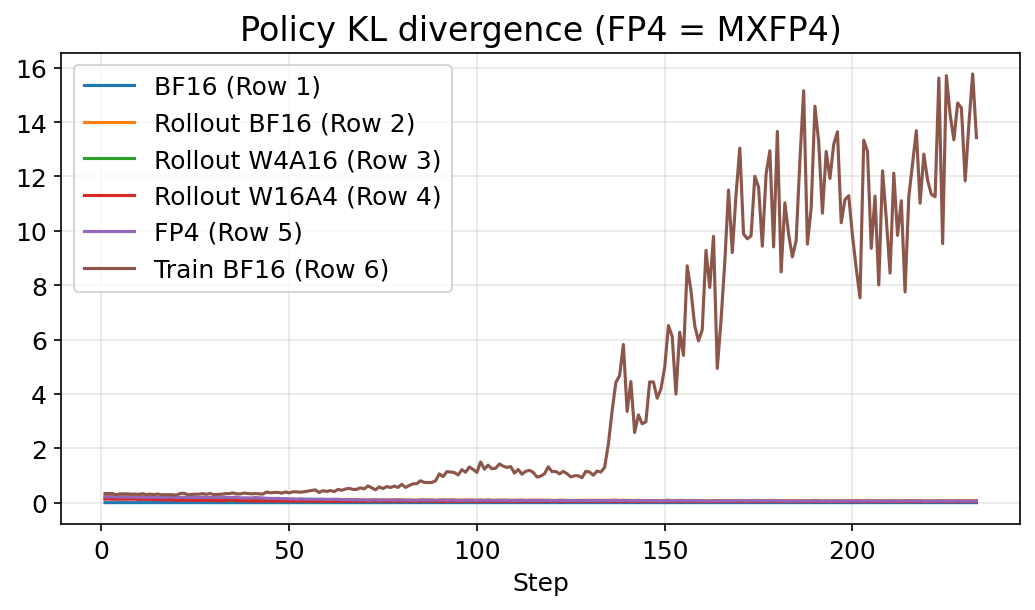}
        \caption{MXFP4}
        \label{fig:second-fig}
    \end{subfigure}
    \caption{Policy KL divergence across differen precision configurations for training and rollout for HiF4 and MXFP4.}
    \label{fig:kl-div}
\end{figure}

Figure~\ref{fig:kl-div} shows the policy KL divergence curves of the six experiments discussed in Section~\ref{sec:issue_fp4_rollout}. We use the following metric to approximate the KL divergence between the training and inference policies,
\begin{align}
    D_{\text{KL}}(\pi_{\text{inference}} \parallel \pi_{\text{train}}) \approx \frac{1}{\sum_t m_t}\sum_t m_t\big(\log\pi_{\text{inference}}(a_t) - \log\pi_{\text{train}}(a_t)\big)
\end{align}
where $\pi_{\text{inference}}$ is the inference policy, $\pi_{\text{train}}$ is the training policy, and $m_t$ is the response mask of token $a_t$. Throughout the learning, training in BF16 exhibits significantly higher KL divergence between the polices than the other five settings, revealing that the performance collapse is associated with policy mismatch. Moreover, for HiF4, the mismatch keeps amplifying as the training continues for BF16-train, reaching a peak KL divergence of 4.157 towards the end. 
In comparison, the HiF4 baseline just peaks at 0.0812 and BF16 baseline peaks at 0.000721. All other runs in Table 1 never exceeds 0.13. Similarly, for MXFP4, BF16-train + FP4-rollout reaches 15.76 but all other runs never exceed 0.5. 

\end{document}